\pdfoutput=1

\documentclass[11pt]{article}

\usepackage{acl}
\usepackage{times}
\usepackage{latexsym}
\usepackage{natbib}
\usepackage{float}
\usepackage{bbding}

\usepackage{amsmath}

\usepackage[T1]{fontenc}

\usepackage[utf8]{inputenc}

\usepackage{microtype}

\usepackage{inconsolata}

\usepackage{graphicx}

\usepackage{booktabs}

\usepackage{multirow}

\usepackage{makecell}

\title{Text-to-TrajVis: Enabling Trajectory Data Visualizations from Natural Language Questions}

\author{
 \textbf{Tian Bai\textsuperscript{1}},
 \textbf{Huiyan Ying\textsuperscript{1}},
 \textbf{Kailong Suo\textsuperscript{1}},
 \textbf{Junqiu Wei\textsuperscript{2}},
 \textbf{Tao Fan\textsuperscript{3}},
 \textbf{Yuanfeng Song\textsuperscript{3}}
\\
\\
 \textsuperscript{1}Jilin University,
 \textsuperscript{2}Macau University of Science and Technology,
 \textsuperscript{3}WeBank AI
}

\begin{document}
\maketitle
\begin{abstract}
This paper introduces the \textbf{Text-to-TrajVis} task, which aims to transform natural language questions into trajectory data visualizations, facilitating the development of natural language interfaces for trajectory visualization systems.
As this is a novel task, there is currently no relevant dataset available in the community. To address this gap, we first devised a new visualization language called Trajectory Visualization Language (TVL) to facilitate querying trajectory data and generating visualizations. Building on this foundation, we further proposed a dataset construction method that integrates Large Language Models (LLMs) with human efforts to create high-quality data. Specifically, we first generate TVLs using a comprehensive and systematic process, and then label each TVL with corresponding natural language questions using LLMs. 
This process results in the creation of the first large-scale Text-to-TrajVis dataset, named \textbf{TrajVL}, which contains 18,140 \textit{(question, TVL)} pairs. Based on this dataset, we systematically evaluated the performance of multiple LLMs (GPT, Qwen, Llama, etc.) on this task. 
The experimental results demonstrate that this task is both feasible and highly challenging and merits further exploration within the research community.
\end{abstract}

\section{Introduction}

The surge in human-generated trajectory data, especially from vehicles, such as the GeoLife+ \cite{amiri2024geolife+}, T-Drive \cite{Yuan2010TdriveDD}, and WorldTrace \cite{Zhu2024UniTrajLA} datasets, has provided researchers with a wealth of spatio-temporal information. Visualization techniques are becoming increasingly important for understanding trajectory patterns and for in-depth analysis in areas such as human mobility pattern analysis \cite{toch2019analyzing,Zhu2024AGO}, infectious disease transmission modeling \cite{amiri2024urban,kohn2023epipol} and animal migration tracking \cite{konzack2019visual}. However, existing query and visualization methods often require specialized knowledge, preventing non-expert users from effectively deriving insights from trajectory data.

Natural Language to Visualization (NL2VIS) technology focuses on the automated translation of natural language queries into visualization specifications \cite{song2024marrying, song2022rgvisnet}. This technology empowers non-expert users to readily visualize trajectory data, offering novel avenues for addressing the aforementioned challenges. To this end, we introduce the Text-to-TrajVis task. In this framework, users pose trajectory data visualization requests using natural language, and the Text-to-TrajVis technology parses these queries to identify key information, including area, time, and relevant query parameters. Subsequently, the natural language query is transformed into Trajectory Visualization Language (TVL), which is then employed to automatically generate a visualization program that renders the queried data as a map or chart. For example, a natural language query that requests trajectory visualization is transformed into a map-based visualization, as illustrated in Figure 1 (see A.1 for the transformation process of other visualization charts).

\begin{figure}[t!]
	\centering
	\includegraphics[width=\columnwidth]{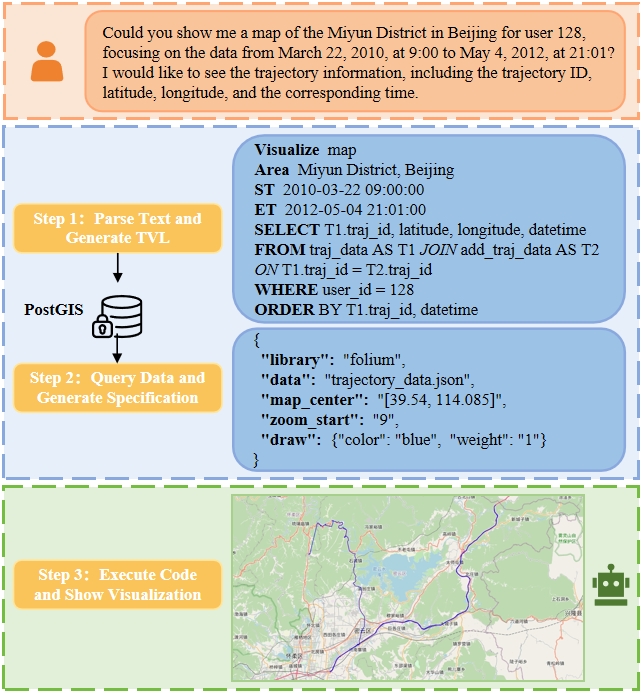} %
	\caption{The process of transforming natural language question into visualization.}
    \vspace{-15pt}
	\label{fig:1}
\end{figure}

Large language models have shown remarkable generalization ability in the field of natural language processing. In addition to making breakthroughs in traditional tasks such as text generation and Q\&A systems, LLMs have enabled the construction of knowledge indexing libraries through Retrieval-Augmented Generation (RAG), significantly improving the accuracy and domain adaptability of the generation tasks. A typical LLM-based workflow for generating trajectory data visualizations involves assembling natural language queries and corresponding examples into prompts, and then prompting the LLM to generate a corresponding TVL. However, to the best of our knowledge, a dedicated NL2VIS dataset for trajectory visualization, suitable for benchmarking LLMs, remains unavailable.

To address this gap, we introduce TrajVL, the first benchmark dataset specifically designed for the Text-to-TrajVis domain. We constructed a generalized template and a systematic methodology for generating TVL. This methodology contains two key stages: First, representative TVLs were generated as seed dataset by incorporating geographical area and temporal information into the template. Second, we developed a tree structure based on query constraints, where each leaf node represents a specific constraint or a combination thereof. Then, constraints added to the TVLs to generate a richer dataset. Recognizing that manual verification requires less time and effort than manual natural language question writing, we used LLMs to generate multiple, distinct natural language questions for each TVL. Each generated natural language question was manually reviewed to ensure it accurately and comprehensively described the corresponding TVL. Our benchmark dataset contains 6,988 TVLs and 18,140 \textit{(question, TVL)} pairs.

We evaluated the proposed TrajVL dataset using various LLMs (e.g., GPT, Qwen, Llama, and DeepSeek). Surprisingly, the best-performing model achieved 74.61\% TVL accuracy in normal scenarios. However, in the scenarios that test the model's spatio-temporal reasoning ability, the best-performing model only achieved about 57.88\% and 68.63\% TVL accuracy in few-shot scenarios. The results of the evaluation indicate that the challenge of the Text-to-TrajVis task deserves further exploration by the research community.

In summary, the main contributions of this paper are as follows:

\begin{itemize}
\item[$\bullet$] We introduce the Text-to-TrajVis task and propose a dataset construction method that combines LLMs with human efforts.
\item[$\bullet$] We introduce TrajVL, the first benchmark dataset designed specifically for the Text-to-TrajVis domain, containing 18,140 \textit{(question, TVL)} pairs.
\item[$\bullet$] We evaluated various LLM-based methods (GPT, Qwen, Llama, etc.) on the dataset to establish the baseline performance and characterize the key properties of the proposed Text-to-TrajVis task.
\end{itemize} 

\section{Related Work}

\subsection{Trajectory Data Storage}
Existing geospatial database systems, such as GeoMesa \cite{hughes2015geomesa}, ArcGIS \cite{law2019getting}, PostGIS \cite{obe2021postgis}, and Google BigQuery \cite{mucchetti2020bigquery}, can be extended to manage trajectory data. PostGIS, in particular, offers a comprehensive set of functions and indexes for storing and querying trajectory data \cite{zhang2010management}. It also supports spatial indexing methods, such as R-trees, which significantly accelerate trajectory queries. The integration of PostGIS with MobilityDB \cite{zimanyi2020mobilitydb} extends its ability to handle large-scale trajectory data. Furthermore, PostGIS provides a rich collection of spatial functions \cite{obe2021postgis} for calculating distances, areas, and lengths, as well as for determining spatial relationships.

\subsection{Trajectory Data Visualization}
Existing trajectory data visualization methods exhibit considerable diversity. One of the most fundamental approaches involves plotting trajectories as line segments on a map \cite{he2019diverse}, which offers a straightforward representation of movement paths. In the trajectory data visualization domain, a variety of specialized languages have been developed for describing, querying, and manipulating trajectory data. GeoJSON \cite{butler2016geojson}, an open standard JSON-based format, is widely used for representing various geospatial data structures. Programming languages and their corresponding libraries, such as Python and JavaScript, offer more flexible programming interfaces for visualizing trajectory data. For instance, the Folium \cite{suma2024geospatial} library in Python offers the functionality of constructing interactive web maps using Leaflet \cite{crickard2014leaflet}. Libraries such as GeoPandas \cite{jordahl2021geopandas}, Shapely \cite{gillies2023shapely}, and Cartopy \cite{elson2022scitools} provide functions for processing and visualizing trajectory data.

\begin{figure}[t!]
	\centering
	\includegraphics[width=\columnwidth]{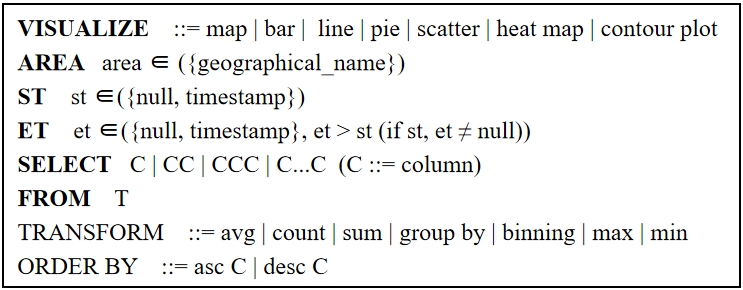} %
	\caption{The structure of Trajectory Visualization Language.}
	\label{fig:2}
\end{figure}

\begin{figure}[t!]
	\centering
	\includegraphics[width=\columnwidth]{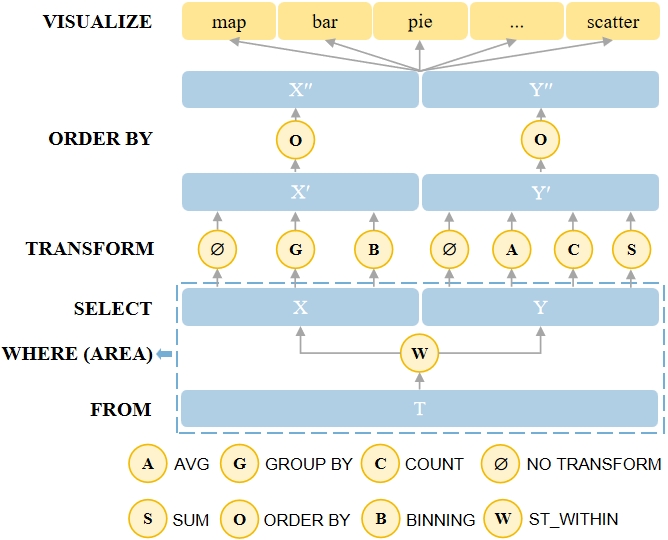} %
	\caption{The visualization workflow for Trajectory Visualization Language.}
    \vspace{-15pt}
	\label{fig:3}
\end{figure}

\subsection{Text-to-Visualization} 
The field of NL2VIS has garnered increasing attention with advancements in visualization technology \cite{cox2001multi,gao2015datatone,liu2021advisor,narechania2020nl4dv,luo2021natural,luo2021synthesizing}. Many studies employ deep neural networks to learn the mapping between NLQs and visualization specifications \cite{dibia2019data2vis,song2022rgvisnet,liu2023multi,li2023graphix,xiang2023g3r}. To facilitate text-to-visualization development, researchers have also introduced corresponding datasets. Luo et al. proposed the NVBench dataset \cite{luo2021synthesizing} as a benchmark for NL2VIS research. Song et al. introduced Dial-NVBench \cite{song2024marrying}, a benchmark dataset designed for interactive visualization research. Currently, LLM-based approaches have become the dominant paradigm in NL2VIS research\cite{gan2021natural,dong2023c3,li2023resdsql,pourreza2023din,gao2023text}, significantly enhancing task performance through the synergistic interplay of semantic parsing and code generation. Pre-trained LLMs have also been increasingly applied to this tasks. For instance, Chat2vis \cite{maddigan2023chat2vis} utilizes a bimodal prompting strategy involving “natural language query + tabular element”, prompting engineering-driven LLMs to generate visualization code. Tian’s team proposed a task decoupling strategy that focuses on visualization task-oriented fine-tuning of LLMs, fine-tuning the FLAN-T5 model \cite{chung2024scaling} to ChartGPT \cite{tian2024chartgpt} in stages.

\begin{table}
\caption{TrajVL dataset statistics}
\label{tab_fwsc}
\centering
\begin{tabular}{p{0.3\linewidth} p{0.2\linewidth} p{0.3\linewidth}}
\toprule
Vis-Type & TVL & (TVL, NLQ)\\
\midrule
Map & 4,873 & 12,377 \\
Bar Chart & 788 & 2,256\\
Line Chart & 727 & 1,997\\
Pie Chart & 600 & 1,510\\
\midrule
Total & 6,988 & 18,140\\
\bottomrule
\end{tabular}
\end{table}

While significant progress has been made in NL2VIS research, trajectory visualization remains largely unexplored. Our work is the first systematic effort to address this research gap. We introduce a specialized dataset (i.e., TrajVis), evaluation metrics, and comprehensive evaluations tailored for this task. Experimental and human evaluations validate the utility of this dataset, and we anticipate that it will facilitate future research in trajectory-aware visual analytics and advance other domain-specific work in NL2VIS applications.

\begin{figure*}[t!]
	\centering
	\includegraphics[width=1\textwidth]{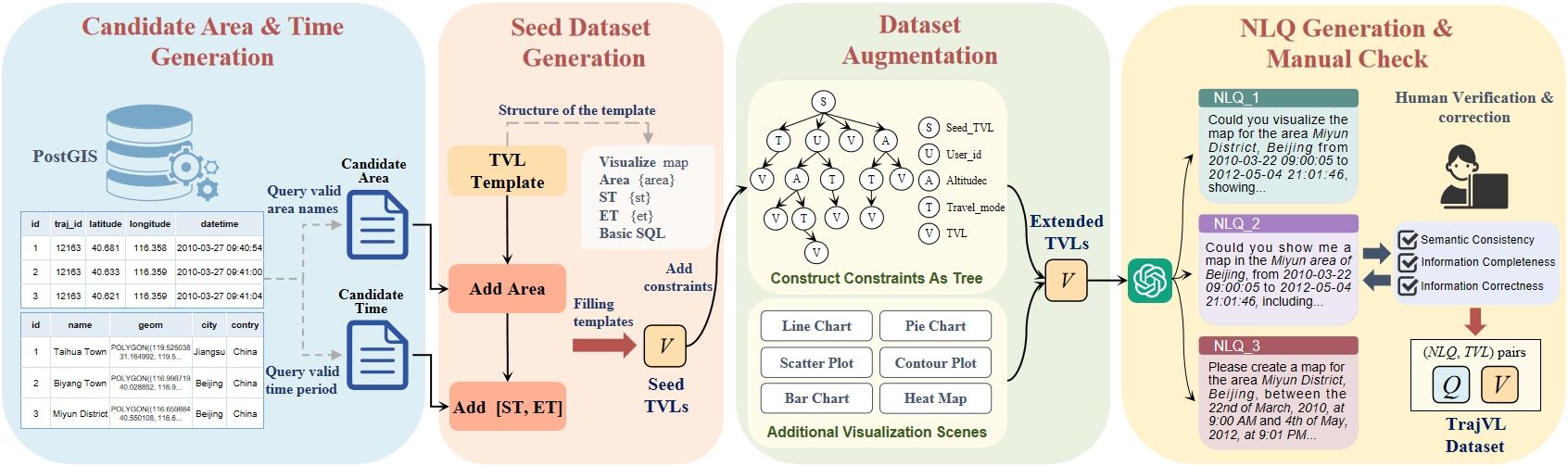} %
	\caption{The pipeline of constructing the TrajVL dataset. The construction of TrajVL involves four main steps: (i) collecting all geographic areas and their associated time ranges from the trajectory database as candidates; (ii) generating representative TVLs as seed data by integrating geographic areas and temporal information into the template; (iii) enriching the dataset by expanding TVLs using a tree-structured augmentation approach and by adding other visualization scenes; and (iv) generating multiple types of natural language questions via LLMs, followed by manual validation.}
    \vspace{-5pt}
	\label{fig:4}
\end{figure*}

\section{TrajVL Datasets}

\subsection{Data Collection \& Preparation}
Trajectory data used for constructing TrajVL are sourced from the GPS trajectory dataset released by the GeoLife project \cite{2011Geolife}. This dataset, which was collected from 182 users from April 2007 to August 2012, contains 17,621 trajectories, each represented by a series of timestamped points. Each point contains latitude, longitude, and altitude information. The dataset also includes a label file for each user, which indicates travel mode labels. User movement trajectories are mainly distributed across more than 30 cities in China, with the majority of the data originating from Beijing.

To obtain information about the administrative boundaries of Chinese provinces at various levels, we utilized the Chinese area boundary data files provided by OpenStreetMap \cite{haklay2008openstreetmap}. This area boundary information was subsequently imported and stored within the PostGIS database. 

\textbf{Trajectory Visualization Language}
We designed a template for generating TVL queries. This template encompasses visualization types, geographical areas, temporal information, and fundamental SQL structures. The \textit{VISUALIZE} component specifies the desired visualization type and transforms the queried data into a visualization query expressed in a specific language (e.g., Python, Vega-Lite). For trajectory data, spatio-temporal information is crucial and can be utilized to impose conditional constraints on the queried data. The spatial dimension enables the specification of geographical entity names via the \textit{AREA} parameter. The temporal dimension employs a dual-anchor constraint mechanism, where start time and end time define a closed interval [\textit{ST, ET}]. Within the SQL structure, the \textit{TRANSFORM} operator integrates trajectory analysis-specific functions, including temporal binning, spatial aggregation (\textit{group by}), and statistical computation (\textit{avg/count/sum/max/min}), to construct a transformation pipeline \textit{T(x) → x'}. The \textit{ORDER BY} extension facilitates a sorting strategy based on the transformed variables (e.g., binning interval, aggregated values), and specifies the sorting direction via \textit{ASC} or \textit{DESC}. Figure 2 defines the structure of the TVL.

The workflow of TVL embeds spatio-temporal semantics into structured queries through a formal constraint mechanism. This mechanism automatically translates the \textit{AREA} and [\textit{ST, ET}] parameters into SQL \textit{WHERE} clause constraints during the syntactic parsing phase. For spatial constraints, the expression \textit{"ST\_Within(geometry\_column, spatial\_filter)"} is dynamically generated. Here, the \textit{spatial\_filter} is parsed as a geofence in WKT format using a library of predefined names. For temporal constraints, the filter condition \textit{"datetime\_column BETWEEN ST AND ET"} is generated (see A.3 for details). The workflow of TVL ensures seamless integration of spatio-temporal constraints with standard SQL syntax, as illustrated in Figure 3.

\subsection{Dataset Construction}

\begin{table*}
\centering
\label{tab_fwsc}
\begin{tabular*}{\linewidth}{@{}lcccccc@{}}
\toprule
Dataset & Size & \makecell{Query \\ Language} & \makecell{Output \\ Type} & Task Type & Construction  & \makecell{Multi-Type \\ NL Question}\\
\midrule
Spider & 10,181 & SQL & Data & NL2SQL & Manual & \XSolidBrush \\
nvBench & 25,750 & VQL & Vis & NL2VIS & Manual+Synthesis & \XSolidBrush\\
GeoQuery & 877 & SQL & Geo-data & NL2GIS & Manual & \XSolidBrush\\
NLMAP V2 & 28,609 & MRL & Geo-data & NL2GIS & Manual & \XSolidBrush\\
OverpassQL & 8,352 & OverpassQL & Geo-data & NL2OSM & Manual & \XSolidBrush\\
\midrule
TrajVL & 18,140 & \textbf{TVL} & \textbf{Geo-vis} & \textbf{NL2TrajVis} & \textbf{Manual+LLM} & \Checkmark\\ 

\bottomrule
\end{tabular*}
\caption{Comparison of TrajVL with other existing Text-to-Query datasets. Notice that Multi-Type NL Question refers to whether each Query supports multiple types of NL questions, not just the entire dataset.}
\vspace{-10pt}
\end{table*}

\textbf{Seed Dataset Generation}
Since the trajectory data is stored in a database, we first collected all geographical areas within the trajectory data to validate areas before incorporating them into the template. These collected areas served as a candidate selections. Next, we extracted the earliest and latest timestamps of trajectories within each area to define temporal ranges. From each range, we selected 3-5 time intervals to form temporal candidates. Finally, representative TVL queries were then generated by selecting geographical areas and time intervals from the candidates.

\textbf{Dataset Augmentation}
We designed a tree structure with additional constraints for trajectory data querying, aiming to enrich and diversify TVL. Besides latitude and longitude in the basic table, the database has auxiliary tables storing extra trajectory attributes. We extracted and organized attributes like user ID, travel mode, and altitude to form query constraints, using them as the foundation for the tree structure where leaf nodes represent specific or combined constraints. By embedding these nodes into the SQL query structure of TVL, we generated a more comprehensive dataset, and the constraints can be optimized as more attributes are added to meet diverse user needs.

We found that complementary trajectory features, such as travel mode and altitude, can be utilized to construct novel visual query paradigms. Therefore, variations in travel mode and altitude within trajectories were selected as core analytical dimensions and integrated with spatio-temporal constraints to generate TVL. This methodology was employed to generate queries for three common visualization types: pie charts, line charts, and bar charts to expand the dataset. This dataset can be extended to support other visualization types, including scatter plots and heat maps. Complete implementation details are provided in A.2.

\textbf{Natural Language Questions Generation}
Considering the labor-intensive and time-consuming process of manual generation of natural language questions, we opted to use LLMs for this task. Considering the paramount importance of area and time information within trajectory data, we sequentially provided three distinct prompts to the LLM to generate diversified descriptions of each TVL, primarily differing in the articulation of geographical areas and times. This approach facilitated the generation of 2-3 corresponding natural language queries for each TVL query. Detailed prompts and generated samples are provided in A.4.

\textbf{Manual Check}
We used a manual evaluation method to assess the quality of the NLQs generated by LLM based on TVL, mainly examining quality indicators in three dimensions: semantic consistency, information completeness and information correctness. The manual assessment results show that the following quality issues exist in the samples: 11.96\% of the samples present semantic redundancy, 13.1\% of the samples have missing information problems, and 8.4\% of the samples contain information errors. Specifically, semantic redundancy refers to the fact that the generated NLQ contains information content that is not related to TVL. Information missing is manifested as the NLQ failing to cover all the necessary fields in the TVL. Information error is manifested by the NLQ's description of key information such as time, area, and SQL being inconsistent with the TVL. To address the above three types of data quality problems, we designed a differentiated prompt strategy (see Appendix A.5 for details) to guide LLMs to make NLQs corrections. The corrected NLQs underwent a second round of manual validation focused on verifying the accuracy of the corrections. Through the iterative correction and validation process, we finally achieved a comprehensive optimization of data quality, ensuring the accuracy and reliability of the generated data. The dataset generation workflow is illustrated in Figure 4.

\subsection{Dataset Analysis}
The dataset contains four prevalent visualization types, including maps, bar charts, line charts, and pie charts. It contains a total of 6,988 visualization instances. For each visualization, TrajVL contains one to several natural language questions, containing a total of 18,140 \textit{(question, TVL)} pairs (Table 1). The percentage of maps in the dataset is 69.7\%, because map is the most common chart type in trajectory data visualization. In addition, visual queries also include those like “Display the percentage of travel by each mode in Beijing over the past year” and “Display the change in altitude of the user's trajectory over a period of time”, etc. Therefore, we expanded the number of three types of visualization charts: bar charts, line charts, and pie charts. These expansions accounted for 11.3\%, 10.4\%, and 8.6\% respectively of the dataset. Our dataset covers 9 provinces and includes a total of 638 areas, most of which are located in Hebei and Beijing.

\begin{table*}[h!]
\centering
\label{tab_fwsc}
\begin{tabular*}{\linewidth}{@{}p{0.085\linewidth}p{0.195\linewidth}p{0.09\linewidth}p{0.105\linewidth}p{0.085\linewidth}p{0.085\linewidth}p{0.085\linewidth}p{0.085\linewidth}@{}}
\toprule
\multirow{2}*{Test Set} & \multirow{2}*{Model} & \multirow{2}*{Vis.Acc} & \multirow{2}*{Axis.Acc} & \multicolumn{4}{c}{Data.Acc} \\
    \cline{5-8}
    \multicolumn{4}{c}{~} & Area & Time & SQL & TVL \\
\midrule
\multirow{4}*{Normal}
& DeepSeek-llm-7b & 99.95 & 86.47 & 82.51 & 81.96 & 40.14 & 35.94\\
~ & Llama3-8b & 99.70 & 89.20 & 86.92 & 94.88 & 57.27 & 53.07\\
~ & Qwen2.5-7b& 99.95 & 98.78 & 98.43 & 99.75 & 64.93 & 64.06\\
~ & GPT-4o-mini & 99.85 & 98.94 & 98.78 & 99.19 & 65.64 & \textbf{64.67}\\
\midrule
\multirow{4}*{Area}
& DeepSeek-llm-7b & 100.0 & 72.93 & 61.94 & 84.64 & 37.56 & 24.33\\
~ & Llama3-8b & 99.70 & 75.92 & 64.88 & 93.87 & 53.22 & 37.61\\
~ & Qwen2.5-7b& 100.0 & 83.53 & 74.96 & 99.65 & 62.75 & 47.24\\
~ & GPT-4o-mini & 99.75 & 82.11 & 72.58 & 98.88 & 64.62 & \textbf{47.74}\\
\midrule
\multirow{4}*{Time}
 & DeepSeek-llm-7b & 100.0 & 84.24 & 79.02 & 75.06 & 37.91 & 27.88\\
~ & Llama3-8b & 99.49 & 88.29 & 84.79 & 89.31 & 50.13 & 41.92\\
~ & Qwen2.5-7b & 100.0 & 97.31 & 95.84 & 93.72 & 62.70 & 55.90\\
~ & GPT-4o-mini & 99.95 & 97.62 & 96.00 & 94.68 & 63.30 & \textbf{56.66}\\
\bottomrule
\end{tabular*}
\caption{Comparison of performance for each LLM on TrajVL.}
\end{table*}

\begin{table*}[h!]
\centering
\label{tab_fwsc}
\begin{tabular*}{\linewidth}{@{}p{0.085\linewidth}p{0.195\linewidth}p{0.09\linewidth}p{0.105\linewidth}p{0.085\linewidth}p{0.085\linewidth}p{0.085\linewidth}p{0.085\linewidth}@{}}
\toprule
\multirow{2}*{Test Set} & \multirow{2}*{Model} & \multirow{2}*{Vis.Acc} & \multirow{2}*{Axis.Acc} & \multicolumn{4}{c}{Data.Acc} \\
    \cline{5-8}
    \multicolumn{4}{c}{~} & Area & Time & SQL & TVL \\
\midrule
\multirow{4}*{Normal} 
& DeepSeek-llm-7b & 98.88 & 86.16 & 81.60 & 96.40 & 65.89 & 58.19\\
~ & Llama3-8b & 99.95 & 96.05 & 94.58 & 97.77 & 74.76 & 71.31\\
~ & Qwen2.5-7b& 100.0 & 98.68 & 98.07 & 99.95 & 76.03 & 74.56\\
~ & GPT-4o-mini & 99.95 & 99.19 & 98.58 & 99.95 & 75.47 & \textbf{74.61}\\
\midrule
\multirow{4}*{Area} 
& DeepSeek-llm-7b & 99.59 & 83.83 & 74.71 & 96.25 & 64.93 & 49.27\\
~ & Llama3-8b & 99.70 & 78.97 & 69.42 & 96.69 & 70.07 & 52.87\\
~ & Qwen2.5-7b& 99.70 & 85.96 & 78.36 & 99.65 & 73.29 & \textbf{57.88}\\
~ & GPT-4o-mini & 99.65 & 84.79 & 76.08 & 99.70 & 74.56 & 57.32\\
\midrule
\multirow{4}*{Time} 
& DeepSeek-llm-7b & 99.95 & 90.22 & 86.67 & 87.33 & 66.65 & 55.30\\
~ & Llama3-8b & 100.0 & 94.98 & 93.22 & 93.66 & 73.20 & 64.38\\
~ & Qwen2.5-7b& 99.95 & 97.87 & 96.15 & 95.39 & 74.76 & 67.66\\
~ & GPT-4o-mini & 100.0 & 97.87 & 96.30 & 96.30 & 75.06 & \textbf{68.63}\\
\bottomrule
\end{tabular*}
\caption{Comparison of performance for each RAG-enhanced LLM on TrajVL.}
\end{table*}

\subsection{Comparison to Other Datasets}
Since we are the first to provide a dataset for the Text-to-TrajVis task, we conducted comparative analyses with the dataset of related tasks (see Table 2). Geoquery \cite{zelle1996learning} is a small dataset with 877 instances. The natural language queries in this dataset are hand-written to simulate geographic information queries that a user might make, and it queries 937 different geographic facts about the United States. NLMAP V2 \cite{lawrence2018improving} provided queries about OpenStreetMap paired with natural language input. However, these queries are written in a restricted query language called Machine Readable Language (MRL), which is an abstraction of the OverpassQL. OverpassQL \cite{staniek2024text} contains queries in the well-established OverpassQL language with additional features for accessing and retrieving geographic information from the OpenStreetMap database. Compared to our dataset, these datasets focus primarily on querying geographic data rather than data visualization. The Spider \cite{yu2018spider} dataset was designed for the Text-to-SQL task. The NvBench \cite{luo2021synthesizing} dataset was constructed based on Spider and is designed to provide high-quality benchmark data for the NL2VIS task. However, it does not include visualization of trajectory data. In contrast to NvBench, our dataset is designed to support trajectory data querying and visualization, with a construction process that combines LLM and manual annotation. LLM can generate large volumes of data rapidly, and manual annotation ensures data quality.

\section{Experimental Setup}

\textbf{Dataset Splitting} We partitioned the dataset into a training set containing 5,012 instances and a test set containing 1,973 instances. The test set was constructed through stratified sampling to ensure representative coverage of diverse query intents and to mitigate bias in the evaluation of specific patterns. Following the methodology described in Section 3.2, we generated either one (when temporal information was absent) or two different natural language questions for each TVL in the Normal test set. As a result, three distinct test sets were generated: Normal test set, Area test set and Time test set. The Area test set and the Time test set were designed to evaluate the accuracy of the LLM in generating TVLs when processing complex spatio-temporal descriptions.

\textbf{Models} To systematically evaluate the performance of diverse LLMs in the TVL generation task, we selected four representative models, which encompass both proprietary and open-source architectures. The selected models include GPT-4 \cite{achiam2023gpt}, Llama-3 \cite{grattafiori2024llama}, Qwen \cite{yang2024qwen2}, and DeepSeek-LLM \cite{bi2024deepseek}. The experiments adopted a few-shot prompting paradigm, Each input prompt contains a system task instruction, an annotated demonstration example of translating a natural language query to TVL, and the target query for parsing. Details of these models are provided in Appendix B.1.

\textbf{Metrics} To evaluate the performance of LLMs in generating TVL, we adopted \textit{Vis Accuracy, Axis Accuracy}, and \textit{Data Accuracy}, metrics commonly employed in text-to-vis tasks \cite{luo2021synthesizing}. Given the critical role of spatio-temporal information in trajectory data, the SQL query for retrieving such data is synthesized from the area, time, and the fundamental SQL query structure. Consequently, within the Data Accuracy assessment, we employ Area Accuracy and Time Accuracy to evaluate the precision of LLM-generated spatio-temporal information. The detailed definition of these metrics can be found in Appendix B.2.

\section{Results and Discussion}
This section presents a comprehensive evaluation of the performance of LLMs on the dataset. Experiments revealed that Qwen2.5-7b and GPT-4o-mini achieved peak performance with a 6-shot prompting configuration, while Deepseek-llm-7b and Llama3-8b achieved peak performance with a 4-shot and 5-shot prompting configuration, respectively (see C.1 for details). Accordingly, Table 3 presents the performance of LLMs on the three test sets, and Table 4 details the results of RAG-enhanced LLMs.

\textbf{Performance of LLMs on Test Sets} Table 3 presents the TVL accuracy results for various models across different test sets. GPT-4o-mini achieved a TVL accuracy of 64.67\% on the Normal test set. However, its performance decreased to 47.74\% and 56.66\% on the Area and Time test sets, respectively. This suggests that GPT-4o-mini is notably sensitive to complex area descriptions, while temporal constraints have a comparatively smaller impact. DeepSeek-llm-7b, Llama3-8b, and Qwen2.5-7b had lower TVL accuracy in both the Area and Time test sets than in the Normal test set. This indicates that the performance of these models is significantly hindered by both complex regional and temporal descriptions. Qwen2.5-7b's performance was second only to that of the GPT-4o-mini when it came to handling spatial and temporal complexity, but further improvements are needed.

The Vis.Acc approached 100\% for most models, indicating robust performance in visual type recognition. Regarding Axis.Acc, all models exhibited significantly higher performance on the Normal and Time test sets than on the Area test set. This is due to the need for centering the map visualization on a specific area, establishing strong dependencies between the visual component and the spatial area. This strong dependency means complex area descriptions may also impair the accurate identification of visualization components. Similarly, Data.Acc showed a significant decrease in performance when processing the Area and Time test sets, suggesting that complex spatio-temporal descriptions pose a challenge to the models. The SQL accuracy remained relatively stable across most models, even in the presence of complex spatio-temporal descriptions. This stability stems from the experimental design, where the SQL evaluation metrics focused on base SQL queries. Area and time information was integrated into the base SQL after TVL generation, thus mitigating the direct impact of spatio-temporal complexity on SQL accuracy during the experimental validation process (as detailed in Section 3.1). However, DeepSeek-llm-7b, Llama3-8b, and Qwen2.5-7b demonstrated a decrease in SQL accuracy. This decline is likely due to difficulties encountered by these models in parsing NLQs containing complex time and area descriptions. These parsing challenges resulted in a reduction of the number of correctly formatted TVL outputs, thus affecting SQL extraction. Conversely, GPT-4o-mini exhibited more stable performance, resulting in fewer fluctuation in SQL accuracy.

\textbf{Performance of RAG-enhanced LLMs on Test Sets} As presented in Table 4, all RAG-enhanced models surpassed their corresponding base models in both TVL and SQL accuracy. The models’ TVL and SQL generation results across the three test sets were consistent with the trends observed in Table 3. However, despite the RAG enhancement, complex spatial and temporal descriptions remained challenging for the models. In the Area test set, Axis.Acc and Area.Acc generally declined across all models compared to the Normal test set, with GPT-4o-mini and Qwen2.5-7b showing the most significant decrease. While Qwen2.5-7b and GPT-4o-mini show the most significant downward trend, their overall performance is still better than DeepSeek-llm-7b and Llama3-8b, which indicates their superior ability in extracting core semantic elements and processing complex spatial descriptions. Similarly, in the Time test set, model performance on temporal descriptions generally decreased, but not as much as in the Area test set. The consistently high performance of GPT-4o-mini and Qwen2.5-7b in the Time test set further demonstrates their superior capability to handle complex temporal descriptions. Despite the high Time accuracy achieved by GPT-4o-mini and Qwen2.5-7b, temporal complexity still affects the accuracy of TVL generation.

Overall, GPT-4o-mini achieved the highest TVL accuracy across Normal test set and Time test set, demonstrating particular robustness in complex scenarios. The RAG-enhanced Qwen2.5-7b model followed closely, achieved the highest TVL accuracy in the Area test set. DeepSeek-llm-7b showed poorer performance with lower TVL accuracy than Llama3-8b. Although the RAG technique significantly improved the performance of LLMs in TVL generation tasks, especially for basic scenarios, its effectiveness remained limited for complex spatio-temporal semantics.

\section{Conclusion}
This paper introduces TrajVL, the first benchmark dataset for the Text-to-TrajVis task. We designed TVL, a novel visualization language, to facilitate trajectory data querying and visualization generation. A dataset construction process combining LLM capabilities with manual refinement was developed, ensuring high-quality NLQ and TVL queries. Subsequently, we evaluated multiple LLMs on TrajVL, revealing challenges in processing spatio-temporal information for Text-to-TrajVis. Experimental results demonstrate that while Retrieval-Augmented Generation (RAG) improves LLM performance, significant limitations persist in handling complex spatio-temporal information.

Future work will investigate enhancing LLMs' performance in processing Text-to-TrajVis tasks. Specifically, we will focus on improving the performance of end-to-end TVL generation from natural language queries with complex spatio-temporal information. This work contributes to the ongoing advancement of the TrajVis field and inspire further research innovations.

\section{Limitations}
We have proposed a dataset generation framework that combines LLM with human intervention, and constructed the first large-scale Text-to-TrajVis benchmark dataset, named TrajVL, to address the lack of benchmark datasets in this field. However, the trajectory data utilized to construct the dataset mainly came from a single data source (GeoLife project dataset). Although the dataset was effectively expanded through a systematic approach, the insufficient amount of additional information on the trajectory data limited the complexity of the constraint tree. In addition, the Area Test and Time Test cannot comprehensively cover all real-world situations with complex descriptions of spatio-temporal information. Therefore, future research should extend the dataset generation framework to other trajectory datasets for generating more general TVLs. The TrajVL dataset can be further extended by integrating richer additional information about trajectories and constructing more complex constraint trees. More diverse and complex spatio-temporal descriptions of real-world trajectories should also be incorporated into the NLQ generation process to enhance the challenge of the dataset.

\bibliography{ref}

\newpage
\onecolumn

\appendix
\section{More Implementation Details}
\label{sec:appendix}

\subsection{Generating visualization specifications based on TVL}
\begin{figure}[H]
	\centering
	\includegraphics[width=\columnwidth]{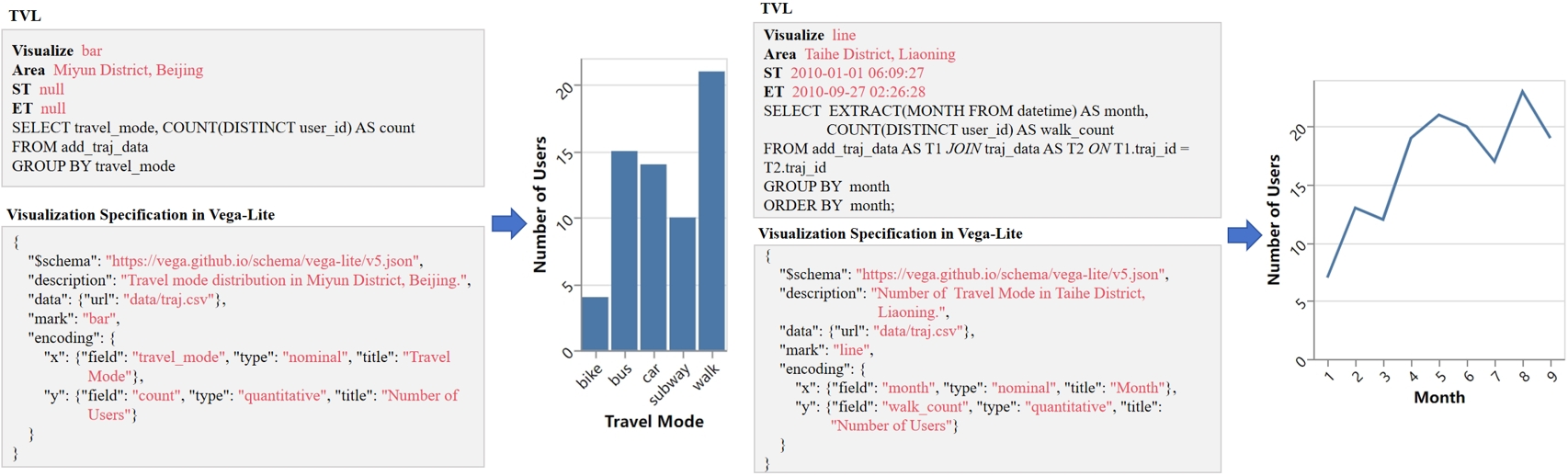} %
	\caption{Examples of the visualization specification in Vega-Lite, and its corresponding visualization.}
	\label{fig:5}
\end{figure}

The proposed Text-to-TrajVis system implements an automatic mapping mechanism from natural language queries to TVLs. As shown in Figure 5, the generated TVL contains three core components: (i) Visualization type statement: specifying the chart presentation form; (ii) Spatio-temporal constraint parameters: constructing geo-spatio-temporal filtering conditions; and (iii) SQL query statement: realizing semantic fusion of trajectory attributes and spatio-temporal coordinates through multi-table association. For standardized chart types (bar charts, line charts, etc.), the system adopts Vega-Lite, a declarative visualization specification framework, which precisely defines the visual coding rules through JSON syntax. The proposed architecture ensures end-to-end interpretability from semantic parsing to visual presentation through formal language transformation mechanisms.

\subsection{Construction Details for Additional Types of Visualization}
\begin{figure}[H]
	\centering
	\includegraphics[width=\columnwidth]{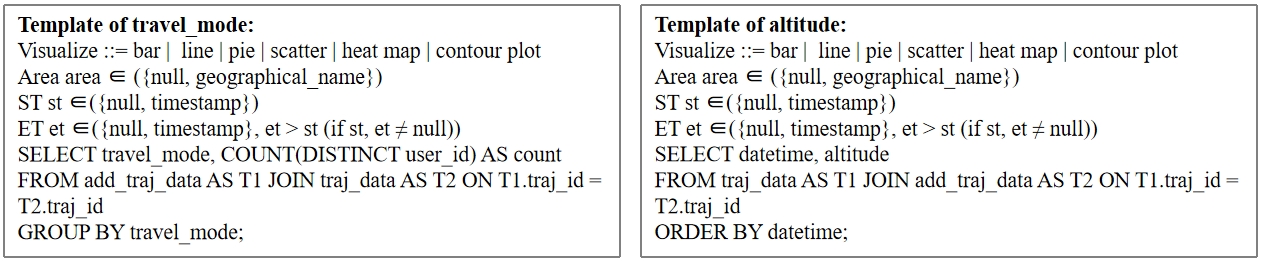} %
	\caption{Templates for extended visualization types built from additional trajectory data.}
	\label{fig:6}
\end{figure}

Query structures are constructed through the integration of spatio-temporal constraints where travel mode and altitude serving as core analytical dimensions (see Figure 6). This approach enables the creation of multiple query scenarios. Specifically, these scenarios support the aggregation of statistical data regarding user distribution across travel modes (e.g., walking, vehicle transportation), thereby enabling comparative visualizations such as bar charts and pie charts. Additionally, the fusion of trajectory metadata and sensor data, achieved through JOIN operations, enables the generation of line graphs that depict variations in altitude. Our current implementation supports three basic chart types: bar charts, line charts, and pie charts. The syntax specification is extensible to more complex visualization types, including heatmaps (e.g., travel mode-based population density) and scatter plots (e.g., location distribution).

\newpage
\subsection{Synthesizing SQL from TVL}
\begin{figure}[H]
	\centering
	\includegraphics[width=\columnwidth]{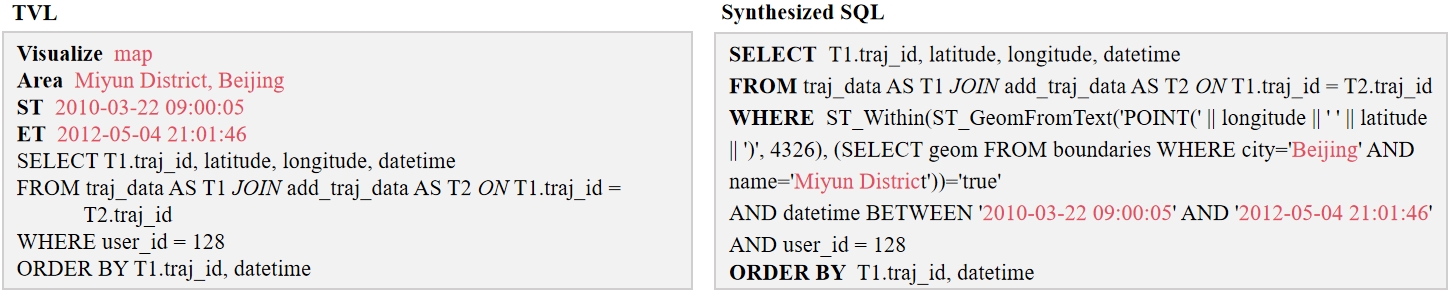} %
	\caption{An example of Synthesizing SQL based on TVL.}
	\label{fig:7}
\end{figure}

Given the TVL specification in Figure 1, the system achieves translation to SQL by parsing the logical structure inherent in TVL. During this conversion, spatio-temporal constraints declared within the TVL are dynamically mapped to the WHERE clause of the resulting SQL query, specifically encompassing the core parameters: Area, ST (Start Time), and ET (End Time). For instance, when TVL specifies that the value of Area is \textit{"Miyun District, Beijing"}, ST is \textit{"2010-03-22 09:00:00"}, ET is \textit{"2012-05-04 21:01:00"}, and the basic SQL structure is "\textit{SELECT user\_id, traj\_id, latitude, longitude, datetime FROM traj\_data ORDER BY user\_id, traj\_id, datetime}". Using these parameters, the system is able to dynamically generate the SQL required to query the data, as shown in Figure 7.

\subsection{Prompts for Generating NLQs}
\begin{figure}[H]
	\centering
	\includegraphics[width=\columnwidth]{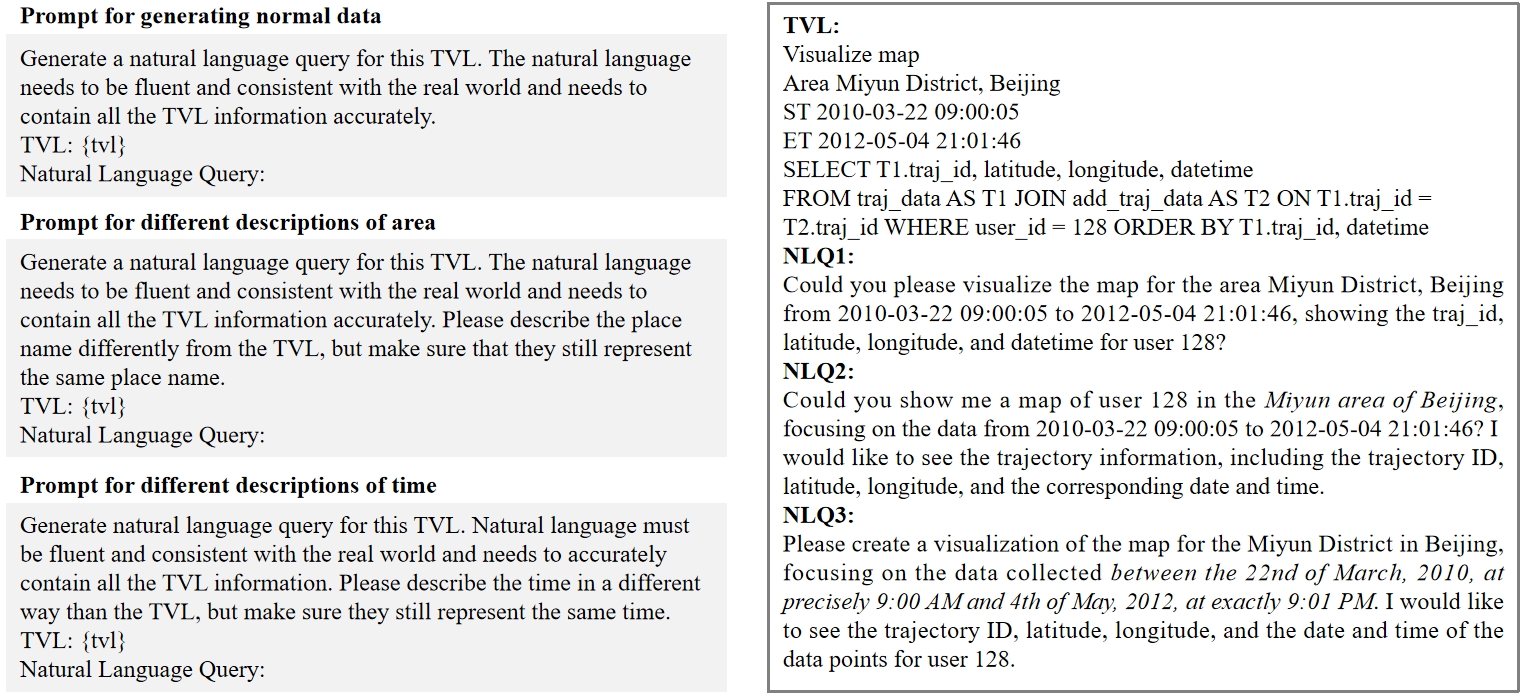} %
	\caption{The prompts for generating NLQs.}
	\label{fig:8}
\end{figure}

We designed prompt templates for LLM. These templates were categorized into three different types: (i) basic data generation: requiring that NLQs maintain semantic accuracy; (ii) diversification of area descriptions: including synonymous substitution of geographic names with enforced consistency of spatial references; and (iii) diversification of temporal descriptions: transforming standard time formats into natural language expressions, preserving the equivalence of the specified time ranges. Through this prompting strategy, the LLM generated three semantically consistent but differently formulated NLQs, as shown in Figure 8.

\subsection{Prompts for Correcting NLQs}
\begin{figure}[H]
	\centering
	\includegraphics[width=\columnwidth]{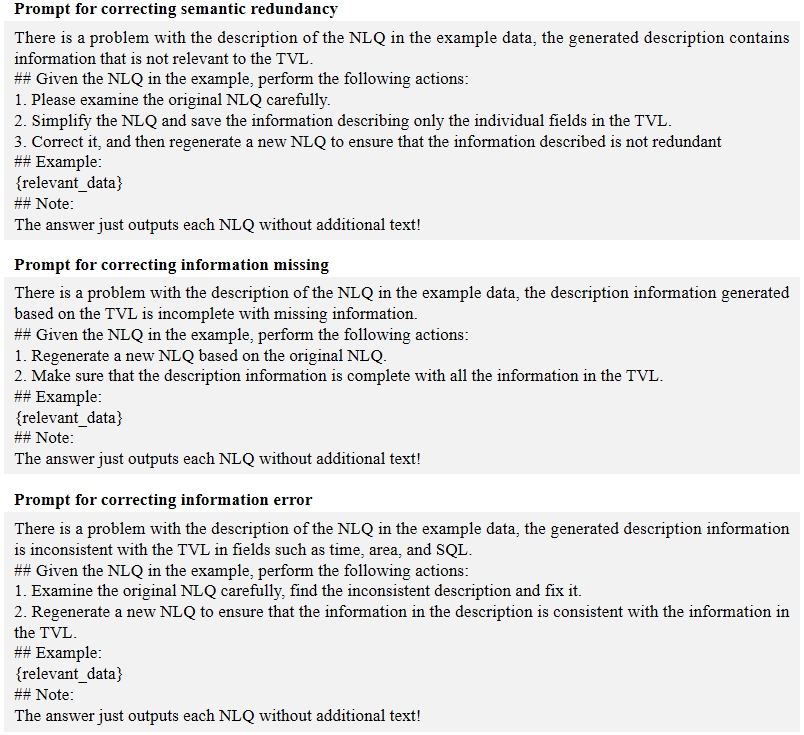} %
	\caption{The prompts for correcting NLQs.}
	\label{fig:8}
\end{figure}

\newpage

\section{More Experimental Setup}

\subsection{Baselines}
To comprehensively evaluate LLMs on the TrajVL dataset, we used a series of prompting strategies (few-shot learning) and RAG techniques. The model configurations are as follows:

\begin{itemize}
\item[$\bullet$] \textbf{Few-shot LLM}: The few-shot prompting is a core mechanism in In-Context Learning, using a limited set of examples to guide LLMs in performing tasks within specialized domains.

\item[$\bullet$] \textbf{RAG for LLM}: Retrieval Augmented Generation (RAG) proposes an alternative methodological paradigm for LLM downstream task support. Unlike direct few-shot prompting, RAG dynamically retrieves the most contextually relevant examples from a curated knowledge base based on the model's inputs. This process enriches the context available for LLM, thus reducing hallucination risks due to insufficient or ambiguous input data.

\item[$\bullet$] \textbf{Llama3-8b}: Llama3-8b, an 8-billion-parameter open-weight language model developed by Meta, is optimized for high efficiency while maintaining robust performance across diverse NLP tasks.  Pretrained on a large-scale multilingual corpus, it demonstrates strong reasoning and text-generation capabilities, positioning it as a versatile choice for applications requiring a balance between model size and inference speed.

\item[$\bullet$] \textbf{Qwen2.5-7b}: Qwen2.5-7b, a 7-billion-parameter large language model developed by Alibaba Group, is engineered to deliver competitive performance in both general and domain-specific tasks.   Trained on diverse multilingual and multimodal corpora, it excels in natural language understanding, generation, and code-related tasks, while maintaining computational efficiency.

\item[$\bullet$] \textbf{DeepSeek-llm-7b}: DeepSeek-llm-7b, a 7-billion-parameter LLM developed by DeepSeek, is designed to provide a strong balance between performance and computational efficiency. Pretrained on extensive text and code corpora, it exhibits competitive capabilities in various natural language understanding and generation tasks, making it a suitable option for applications where resource constraints are a consideration.

\item[$\bullet$] \textbf{GPT-4o-mini}: GPT-4o-mini, a smaller version in OpenAI's GPT-4o series, leverages advanced architectures and training methodologies for strong performance on diverse NLP tasks. Designed for efficiency and responsiveness, it offers a balance between high-quality generation and reduced latency, making it suitable for applications requiring both sophisticated language understanding and rapid response times.
\end{itemize}

\subsection{Metrics}
We use the same metrics as NVBench, including Vis Accuracy, Axis Accuracy, and Data Accuracy, to assess LLM performance in TVL generation. Since spatio-temporal information is important for trajectory data, SQL statements for trajectory queries are synthesized based on area, time and basic SQL query structure. Consequently, Data Accuracy is decomposed into Area Accuracy, Time Accuracy, SQL Accuracy and TVL Accuracy.

The evaluation metrics are formally defined as follows:

\begin{itemize}
\item[$\bullet$] \textbf{Vis Accuracy (Vis.Acc)}: This metric evaluates the accuracy of the Visualize component of the TVL generated by the model, measuring the model's ability to identify the type of visualization. It is calculated as:

\begin{equation*}
Acc_{type} = \frac{N_{type}}{N}
\end{equation*}

Where $ N_{type} $ represents the count of exact match visualization types, $ N $ represents the total number of visualization types in the test set.

\item[$\bullet$] \textbf{Axis Accuracy (Axis.Acc)}: This metric evaluates the performance of the model in recognizing vis components. For map visualizations, it evaluates the accuracy of the Area generated by the model, since proper visualization requires centering of the specified area for map visualizations. For the x-axis and y-axis components of other types of visualizations such as bar, line, pie, etc., we measure the Select component of the vis query. It is calculated as:

\begin{equation*}
Acc_{comp} = \frac{N_{comp}}{N}
\end{equation*}

Where $ N_{comp} $ represents the count of exact matches to the Area or Select component, $ N $ represents the total number of queries in the test set.

\item[$\bullet$] \textbf{Area Accuracy (Area.Acc)}: This metric evaluates the accuracy of the Area component of the TVLs generated by the model to reveal the model's performance in recognizing geographical area names. It is calculated as:

\begin{equation*}
Acc_{area} = \frac{N_{area}}{N}
\end{equation*}

Where $ N_{area} $ represents the count of exact matching Area components, $ N $ represents the total number of queries in the test set.

\item[$\bullet$] \textbf{Time Accuracy (Time.Acc)}: This metric evaluates the accuracy of the Time component of the TVLs generated by the model to reveal the model's performance in recognizing temporal information. It is calculated as:

\begin{equation*}
Acc_{time} = \frac{N_{time}}{N}
\end{equation*}

Where $ N_{time} $ represents the count of exact matching Time components, $ N $ represents the total number of queries in the test set.

\item[$\bullet$] \textbf{SQL Accuracy (SQL.Acc)}: This metric evaluates the accuracy of the model in generating critical SQL by parsing and comparing SQL query structures in TVL. Specifically, it eliminates interference from syntactically equivalent but ordering differences through normalization (e.g., alphabetizing logical conditions in WHERE clauses), and recursively compares SQL structures through parse trees. It is calculated as:

\begin{equation*}
Acc_{sql} = \frac{N_{sql}}{N}
\end{equation*}

Where $ N_{sql} $ represents the count of exact matching SQL queries, $ N $ represents the total number of SQL queries in the test set.

\item[$\bullet$] \textbf{TVL Accuracy (TVL.Acc)}: This metric evaluates the accuracy of the model in generating a complete TVL. Based on the structure of the TVL, a model is considered to be able to generate the TVL correctly when it correctly generates Visualize, Area, Time and SQL. It is calculated as:

\begin{equation*}
Acc_{tvl} = \frac{N_{tvl}}{N}
\end{equation*}

Where $ N_{tvl} $ represents the count of exact match TVLs, $ N $ represents the total number of TVLs in the test set.
\end{itemize}

\subsection{Implementation Details.}
Our experiments used three open-source LLMs (Llama3-8b, Qwen2.5-7b, DeepSeek-llm-7b) and one closed-source LLM (GPT-4o-mini). We conducted experiments in both the baseline configuration and the RAG-enhanced configuration. In the baseline configuration, the LLM was used directly. In the RAG-enhanced configuration, a vector library constructed using the \textit{all-mpnet-base-v2} model was used to provide context prompts. During the inference process, the top-k most similar samples are retrieved from the vector library, where k ranged from 1 to 3. The temperature parameter for all LLMs was set to 0.1 and the max\_length was set to 2048 tokens.

\subsection{Prompt for Experiment }
\begin{figure}[H]
	\centering
	\includegraphics[width=\columnwidth]{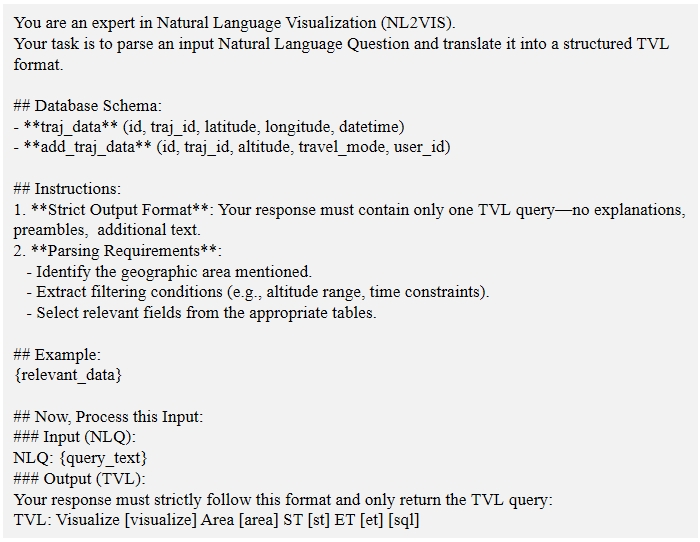} %
	\caption{The prompt for LLMs to generate TVLs in experiment.}
	\label{fig:5}
\end{figure}

\newpage
\section{More Experimental Details}

\subsection{Performance of Few-shot LLMs on Dataset}

\begin{figure}[H]
	\centering
	\includegraphics[width=\columnwidth]{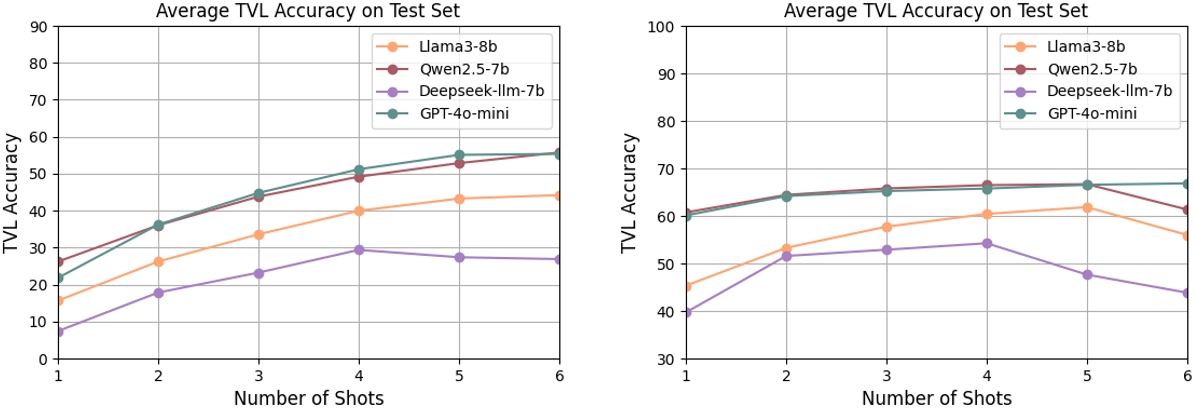} %
	\caption{Average performance of LLMs on three test sets.}
	\label{fig:5}
\end{figure}

We systematically evaluated the impact of few-shot learning on model performance by experimentally comparing the performance of Llama3-8b, Qwen2.5-7b, DeepSeek-llm-7b, and GPT-4o-mini under the baseline configuration and RAG enhancement, as shown in Figure 10. The experimental results show that the average accuracy of LLMs under the base setting shows an increasing trend for both 1-5 shot scenarios. The performance of DeepSeek-llm-7b starts to decrease at 4-shot. The average accuracy of RAG-enhanced LLMs shows an increasing trend across 1-5 shot scenarios. The performance of DeepSeek-llm-7b starts to decrease at 4-shot. The performance of Llama3-8b and Qwen2.5-7b starts to decrease at 5-shot. Consequently, we analyzed in detail the experimental results for the sample scenario with the best performance of LLMs for both experimental settings in Section 5.

\newpage
\subsection{One-shot LLMs' Performance Comparison}
\begin{table*}[h!]
\centering
\label{tab_fwsc}
\begin{tabular*}{\linewidth}{@{}p{0.085\linewidth}p{0.195\linewidth}p{0.09\linewidth}p{0.105\linewidth}p{0.085\linewidth}p{0.085\linewidth}p{0.085\linewidth}p{0.085\linewidth}@{}}
\toprule
\multirow{2}*{Test Set} & \multirow{2}*{Model} & \multirow{2}*{Vis.Acc} & \multirow{2}*{Axis.Acc} & \multicolumn{4}{c}{Data.Acc} \\
    \cline{5-8}
    \multicolumn{4}{c}{~} & Area & Time & SQL & TVL \\
\midrule
\multirow{4}*{Normal} 
& DeepSeek-llm-7b & 96.10 & 49.77 & 42.02 & 51.19 & 17.79 & 8.92\\
~ & Llama3-8b & 99.44 & 67.76 & 67.11 & 66.95 & 24.73 & 20.27\\
~ & Qwen2.5-7b & 100.0 & 77.70 & 75.57 & 81.65 & 35.48 & 30.82\\
~ & GPT-4o-mini & 79.02 & 75.82 & 78.16 & 66.19 & 29.19 & 23.01\\
\midrule
\multirow{4}*{Area} 
& DeepSeek-llm-7b & 100.0 & 46.43 & 28.94 & 42.02 & 14.34 & 5.27\\
~ & Llama3-8b & 99.34 & 56.41 & 44.70 & 66.09 & 21.79 & 12.77\\
~ & Qwen2.5-7b & 99.95 & 67.56 & 56.72 & 89.15 & 34.77 & 22.71\\
~ & GPT-4o-mini & 92.85 & 71.41 & 65.99 & 70.05 & 36.19 & 18.80\\
\midrule
\multirow{4}*{Time} 
& DeepSeek-llm-7b & 99.95 & 52.51 & 36.09 & 34.21 & 16.22 & 8.01\\
~ & Llama3-8b & 99.80 & 66.70 & 61.43 & 59.66 & 19.46 & 13.89\\
~ & Qwen2.5-7b & 100.0 & 77.04 & 72.93 & 73.75 & 72.74 & 25.09\\
~ & GPT-4o-mini & 90.47 & 80.54 & 82.32 & 54.74 & 35.28 & 20.58\\
\bottomrule
\end{tabular*}
\caption{Comparison of performance for each LLM on TrajVL.}
\end{table*}

\begin{table*}[h!]
\centering
\label{tab_fwsc}
\begin{tabular*}{\linewidth}{@{}p{0.085\linewidth}p{0.195\linewidth}p{0.09\linewidth}p{0.105\linewidth}p{0.085\linewidth}p{0.085\linewidth}p{0.085\linewidth}p{0.085\linewidth}@{}}
\toprule
\multirow{2}*{Test Set} & \multirow{2}*{Model} & \multirow{2}*{Vis.Acc} & \multirow{2}*{Axis.Acc} & \multicolumn{4}{c}{Data.Acc} \\
    \cline{5-8}
    \multicolumn{4}{c}{~} & Area & Time & SQL & TVL \\
\midrule
\multirow{4}*{Normal} 
& DeepSeek-llm-7b & 100.0 & 82.01 & 70.25 & 88.55 & 57.27 & 46.27\\
~ & Llama3-8b & 99.29 & 82.26 & 76.28 & 84.64 & 61.33 & 51.39\\
~ & Qwen2.5-7b & 100.0 & 96.76 & 95.13 & 98.07 & 73.59 & 70.86\\
~ & GPT-4o-mini & 96.05 & 94.88 & 94.27 & 95.84 & 70.45 & 68.98\\
\midrule
\multirow{4}*{Area}
& DeepSeek-llm-7b & 99.75 & 77.90 & 61.23 & 87.58 & 46.68 & 30.50\\
~ & Llama3-8b & 98.94 & 72.17 & 57.02 & 85.71 & 54.28 & 36.54\\
~ & Qwen2.5-7b & 99.75 & 81.65 & 70.45 & 98.68 & 66.35 & 48.61\\
~ & GPT-4o-mini & 98.68 & 82.41 & 72.53 & 94.98 & 66.65 & 47.80\\
\midrule
\multirow{4}*{Time}
& DeepSeek-llm-7b & 99.90 & 89.25 & 77.60 & 79.07 & 53.93 & 42.32\\
~ & Llama3-8b & 99.09 & 84.90 & 76.23 & 83.73 & 59.45 & 48.05\\
~ & Qwen2.5-7b & 99.95 & 96.86 & 94.58 & 94.48 & 70.70 & 62.75\\
~ & GPT-4o-mini & 98.94 & 95.95 & 94.58 & 94.08 & 70.25 & 63.56\\
\bottomrule
\end{tabular*}
\caption{Comparison of performance for each RAG-enhanced LLM on TrajVL.}
\end{table*}
Multiple LLMs were evaluated on three test sets, including GPT-4o-mini, Llama3-8b, DeepSeek-llm-7b, and Qwen2.5-7b. Table 5 and Table 6 summarize the performance of these LLMs in both baseline and RAG-enhanced configurations under the 1-shot setting. The results indicate that RAG yielded a substantial improvement in TVL accuracy of these models on the three test sets. While RAG-enhanced 1-shot learning approached the performance of multi-shot learning on basic tasks, a performance gap remained in complex scenarios. In particular, in the Area test set, the TVL accuracy in one-shot scenarios is significantly lower than that of few-shot. In normal scenarios, RAG significantly improved semantic parsing and TVL generation capabilities, with DeepSeek-llm-7b showing a notable increase in TVL and SQL generation accuracy. However, the benefits of RAG were limited in complex scenarios. For instance, the TVL accuracy of Qwen2.5-7b in the Area test set increased only from 22.71\% to 48.61\%. This increase, which is much lower than the percentage improvement in normal scenarios, indicates the complex area descriptions that pose a challenge to the models.

We also observed that In-Context Learning (ICL) effectively mitigates performance deficits in specialized domains, although the number of contextual examples significantly influenced model performance. Specifically, in the three experiments detailed in Section 5, the top-performing GPT-4o-mini achieved TVL accuracies of 64.67\%, 47.74\%, and 56.66\% across the three test sets, respectively, demonstrating a substantial improvement over the 1-shot configurations. While RAG-enhanced LLMs in a 1-shot setting did not surpass the best-performing few-shot setups, their performance remained closely competitive, highlighting the importance of external knowledge retrieval through RAG.

\end{document}